\documentclass{article}

\usepackage[preprint, nonatbib]{neurips_2019}

\usepackage[utf8]{inputenc} 
\usepackage[T1]{fontenc}    
\usepackage{hyperref}       
\usepackage{url}            
\usepackage{booktabs}       
\usepackage{amsmath,amsfonts,amssymb,amsthm}
\usepackage{mathtools}
\usepackage{nicefrac}       
\usepackage{microtype}      
\usepackage{gensymb}
\usepackage[dvipsnames]{xcolor}
\usepackage[normalem]{ulem}
\usepackage{graphicx}
\usepackage{hyperref}
\usepackage{float}
\hypersetup{colorlinks=true,linkbordercolor=red,linkcolor=green,pdfborderstyle={/S/U/W 1}}
\newif{\ifhidecomments}
\usepackage{amsmath}

\DeclareMathOperator*{\argmin}{arg\,min}

\title{Reproducing BowNet: Learning Representations by Predicting Bags of Visual Words}

\newcommand*\samethanks[1][\value{footnote}]{\footnotemark[#1]}
\author{%
  Harry Nguyen\thanks{Equal contribution} \\
  Electrical and Computer Engineering \\
  University of Waterloo \\
   \texttt{harryguyen@uwaterloo.ca}
  \And Stone Yun\samethanks \\
  Systems Design Engineering \\
  University of Waterloo \\
   \texttt{s22yun@uwaterloo.ca}
   
  \And Hisham Mohammad\samethanks \\
  Systems Design Engineering\\
  University of Waterloo\\
  \texttt{hgmohammad@uwaterloo.ca} \\

}

\begin{document}

\maketitle

\section*{\centering Reproducibility Summary}

\textit{This work aims to reproduce results from the CVPR 2020 paper by Gidaris et al. Self-supervised learning (SSL) is used to learn feature representations of an image using an unlabeled dataset. This work proposes to use bag-of-words (BoW) deep feature descriptors as a self-supervised learning target to learn robust, deep representations. BowNet is trained to reconstruct the histogram of visual words (ie. the deep BoW descriptor) of a reference image when presented a perturbed version of the image as input. Thus, this method aims to learn perturbation-invariant and context-aware image features that can be useful for few-shot tasks or supervised downstream tasks. In the paper, the author describes BowNet as a network consisting of a convolutional feature extractor $\Phi(\cdot)$ and a Dense-softmax layer $\Omega(\cdot)$ trained to predict BoW features from images. After BoW training, the features of $\Phi$ are used in downstream tasks. For this challenge we were trying to build and train a network that could reproduce the CIFAR-100 accuracy improvements reported in the original paper. However, we were unsuccessful in reproducing an accuracy improvement comparable to what the authors mentioned. This could be for a variety of factors and we believe that time constraints were the primary bottleneck.}

\section*{Introduction Scope of Reproducibility}
\vspace{-0.1in}
Previous works have found some intriguing results when applying KMeans clustering to the learned feature maps of trained CNNs. This paper looks to build on these "Deep Clustering" works and use entirely self-supervised representations to build bag-of-words representations. 

The main claim of \cite{gidaris2020bownet} is that the high-level, self-supervised learning (SSL) representations of convolutional neural networks (CNN) can be discretized into deep bag-of-words (BoW) descriptors that are encoded with enough rich, semantic information that they can be used as a second set of SSL targets for learning context-aware, robust CNN features. The authors first train an initial CNN (in their case, RotNet) with the self-supervised task of rotation prediction. This network’s higher-level feature maps are then discretized with K-means clustering to construct a visual BoW vocabulary. Afterwards, a second, randomly initialized network (referred to as BowNet) is trained to re-create the same corresponding BoW histograms for a given reference image after perturbations have been applied (Ie. the perturbed/noisy image is the input to BowNet). Once trained, the convolutional feature extractor of BowNet can be frozen and used for downstream tasks such as few-shot classification or transfer learning to a supervised learning task. The authors demonstrate that classification finetuning with BowNet leads to significantly better performance than training on RotNet features (around 10\% improvement for CIFAR-100). The authors also demonstrate the value of their method for few-shot learning tasks. We seek to reproduce and verify the results claimed on CIFAR-100 classification. Specifically, the claim that training a linear classifier with a frozen BowNet feature extractor leads to better accuracy than training on feature maps of a frozen RotNet. We will follow the procedure outlined in their implementation and compare the classification accuracy of (Linear Classifier + BowNet descriptors) vs (Linear Classifier + RotNet feature maps). By specifically investigating the claim of superior performance, we can avoid the issues of trying to recreate the author’s exact RotNet implementation and instead focus on the value of BowNet training. Due to constraints on our compute resources, we will use a more compact, basic ResNet-like architecture. Since there is no code available, we will be implementing it ourselves. 

\newpage
\section{Methodology}

The aforementioned implementation of a BowNet feature extractor involves 2 main stages:

\subsection{Rotation Prediction network (RotNet)}

A visual vocabulary/codebook is constructed from clustering the feature maps of an initial feature extractor trained on some self-supervised task. In \cite{gidaris2020bownet}, the authors choose to use rotation prediction as the self-supervised pretraining task and K-Means clustering to generate the codebook. The BoW representations that would be generated from this CNN's clustered feature maps are used as ground truth for the following stage. 

The self-supervised rotation prediction task is chosen due to its demonstrated value in \cite{gidaris2018rotnet} for forcing the CNN to learn contextually relevant, salient representations of an image. A set of rotations, namely, rotations of 0\textdegree, 90\textdegree, 180\textdegree and 270\textdegree are applied to each image in the CIFAR-100 dataset and is sent as input to the network. During the 4-class classification pretext task, the CNN learns semantic image features such as the location, type and pose of objects in the image. After learning contextual representations through this first SSL task, spatially dense, discrete descriptors of the image are made by clustering the feature maps generated from one of the layers. For a chosen feature map, $\hat\Phi(x)$ of depth $C$, $C$-dimensional feature vectors, $\phi_u$ are sampled at each spatial location $u \in U$ where $U$ is the set of all spatial locations in $\hat\Phi(x)$. We generate a representative set of feature vectors by passing every image in the dataset $X$ through the trained RotNet and densely sampling feature vectors as described. This feature vector set is then clustered into $K$ centroids via K-Means clustering. Ie. we are trying to find the set of vectors $V = [v_1, v_2, ..., v_K]$ that minimize the objective in Eq.~\ref{eq:kmeans_cluster}. $V$ forms our visual vocabulary of the feature map $\hat\Phi(x)$. Following the generation of the descriptors, the BoW representation is expressed by binning the frequency of occurrence of the visual words appearing in the image. The resulting vector is converted to a probability distribution via dividing by the L1-norm, resulting in a soft categorical BoW histogram for each input image. We note that moving forward, we follow the convention set out by the authors and refer to the feature extractor of RotNet as $\hat\Phi(x)$ while the feature extractor of BowNet is simply $\Phi(x)$. In more general terms, $\hat\Phi(x)$ is meant to refer to the feature extractor from which the visual vocabulary is being generated and $\Phi(x)$ is the feature extractor of the network that is learning to reproduce the BoW representation. Following~\cite{gidaris2020bownet}, we use a codebook with $K = 2048$ centroids.

\begin{equation}
    \min_{V} \sum_{x \in X} \sum_{u \in U}[\argmin_{k} \lVert{\phi_u - v_k}\rVert_{2}^{2}]
    \label{eq:kmeans_cluster}
\end{equation}

\subsection{Reconstructing BoW histograms (BowNet)}
After training the RotNet feature extractor and generating a visual vocabulary from one of its feature maps, a second SSL task is used to further learn more robust representations. BowNet is trained to predict the histogram of visual words of a given image when a perturbed version of the image is the input to the CNN. The parameters of BowNet are trained against the BoW representation computed from the RotNet feature map extracted from the \textit{unperturbed} version of the image.

The perturbations are as follows: 
\begin{itemize}
\item Color jittering (random changes in brightness, contrast, saturation, and hue of an image)
\item Conversion to grayscale
\item Random image cropping 
\item Distortion in scale or aspect ratio 
\item Horizontal flips
\end{itemize}

We train BowNet to minimize the cross-entropy loss between the predicted BoW representation generated by BowNet from the perturbed image and the actual BoW representation of the unperturbed input image. Specifically, the crossentropy between the output of BowNet's softmax layer and the reference image's BoW representation. The idea being that this SSL task forces BowNet to learn representations that are robust to perturbation. Ie. Contextually relevant, high-level representations that are invariant to low-level image perturbations. 

\subsection{Design Considerations}

Upon reviewing the original RotNet \cite{gidaris2018rotnet} (Gidariss et al) paper we saw that the authors mentioned using the feature maps from the sixth layer of their self-supervised ConvNet to be best for CIFAR-10 training. However, from our understanding, the RotNet trained in the BowNet paper uses the feature maps from the 3rd residual block to train a linear classifier on CIFAR-100 classification, corresponding to the 25th conv layer in the network. Given that the authors mentioned a degradation in classification accuracy when using the later conv feature maps, it is unclear whether we would still observe the same accuracy improvement if RotNet was trained according to the original setup of its paper. Therefore, we also trained a linear classifier on the 7th conv layer of our RotNet architecture (corresponding to the output of Resblock2\_128b in Figure~\ref{fig:rotnet}).

For our RotNet training pipeline, we reused the author's data-generation code for generating the batches of rotated images \cite{rotnetgithub}. However, the main training pipeline for rotation prediction, generating the RotNet-codebook, training BowNet, and training a linear classifier for representation evaluation are implemented from scratch in PyTorch. Overall, the training tasks are all fairly lightweight and do not require intensive hardware requirements. However, we did need to make some modifications to the CNN architecture and KMeans clustering due to our limited hardware. The Wide-Residual-Network-28-10 (WNR-28-10) was much too large to fit in our GPU memory and similarly, KMeans clustering was too compute intensive. For a Conv256 layer with $8\times8$ feature maps, we would have 64 vectors of length 256 associated with a single input image. Scaling this up to account for ${50, 000 \times 4}$ (4 different rotations) leads to a fairly large clustering problem. Therefore, we resorted to using scikit-learn's~\cite{sklearn} Mini-batch K-means implementation for clustering the RotNet feature maps.

\begin{figure}[h!]
\caption{Pipeline from the original paper}
\centering
\includegraphics[scale=0.3]{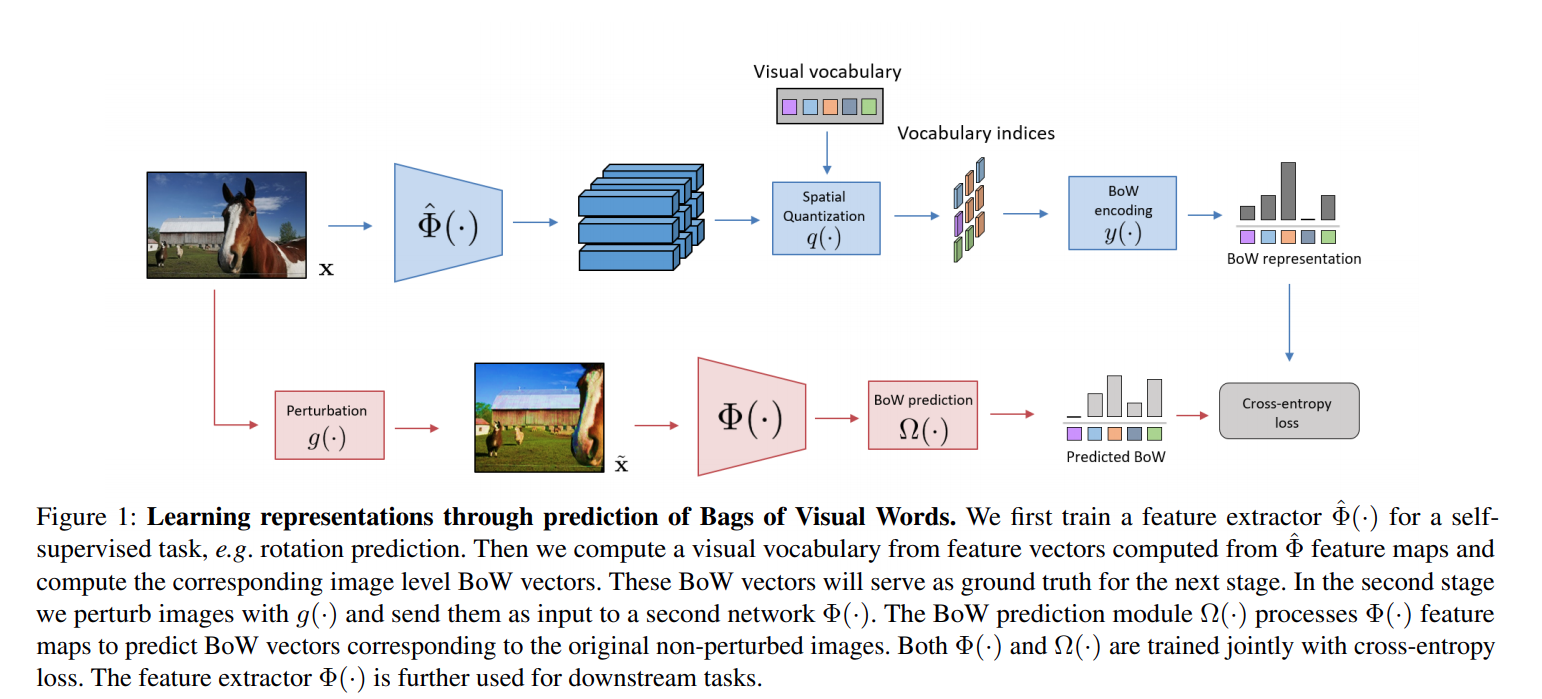}
\label{figure:pipeline}
\end{figure}

\subsection{Model descriptions}
\begin{figure*}[h!]
\centering
\includegraphics[scale=0.7]{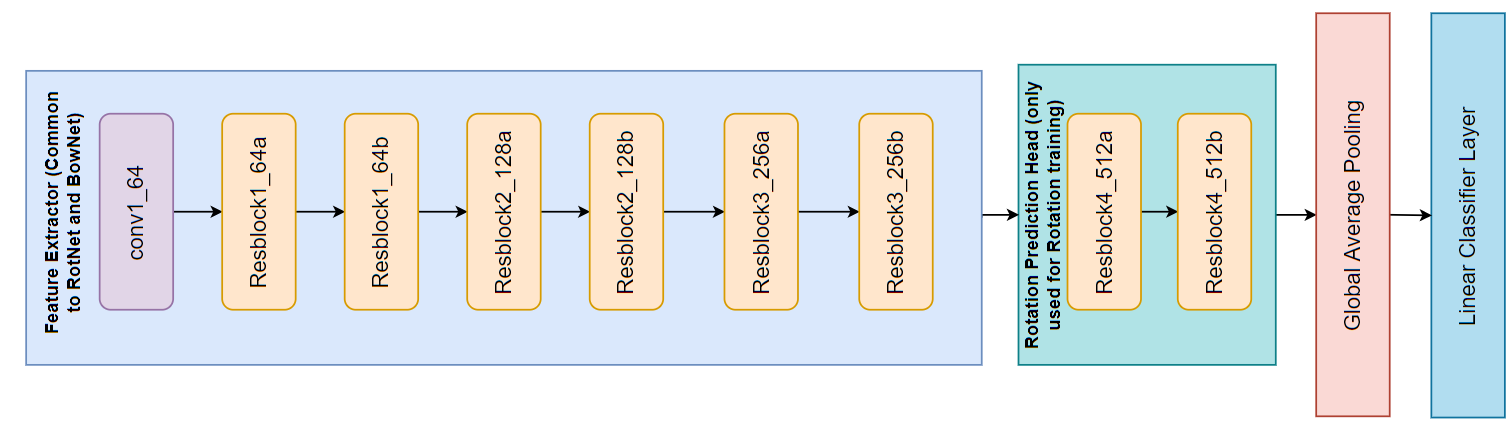}
\caption{Rotation training CNN. This architecture has an extra residual block to act as part of the "rotation prediction head." The authors reported better features being extracted when training RotNet in this setup.}
\label{fig:rotnet}
\end{figure*}

\begin{figure*}[h!]
\centering
\includegraphics[scale=0.7]{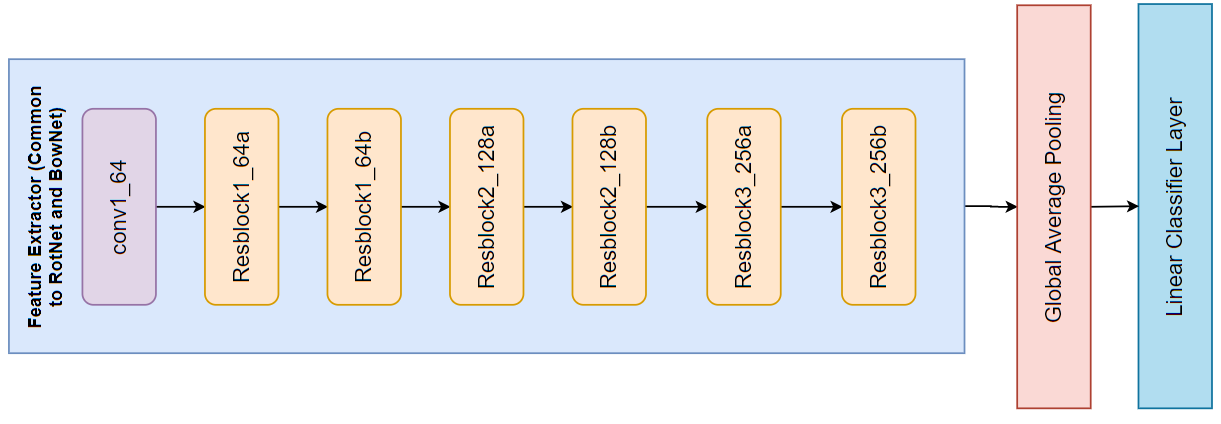}
\caption{BoW reconstruction CNN. In this architecture, the rotation prediction head is discarded and the feature maps of Resblock3\_256b are fed to the reparametrized linear-plus-softmax layer after Global Average Pooling is performed. We train BowNet to minimize the cross-entropy between the BoW representation of the unperturbed image and the Predicted BoW of BowNet (which is just the output of a K-unit softmax layer where K is the number of clusters in our BoW codebook).}
\label{fig:bownet}
\end{figure*}

\begin{figure*}[h!]
\centering
\includegraphics[scale=0.7]{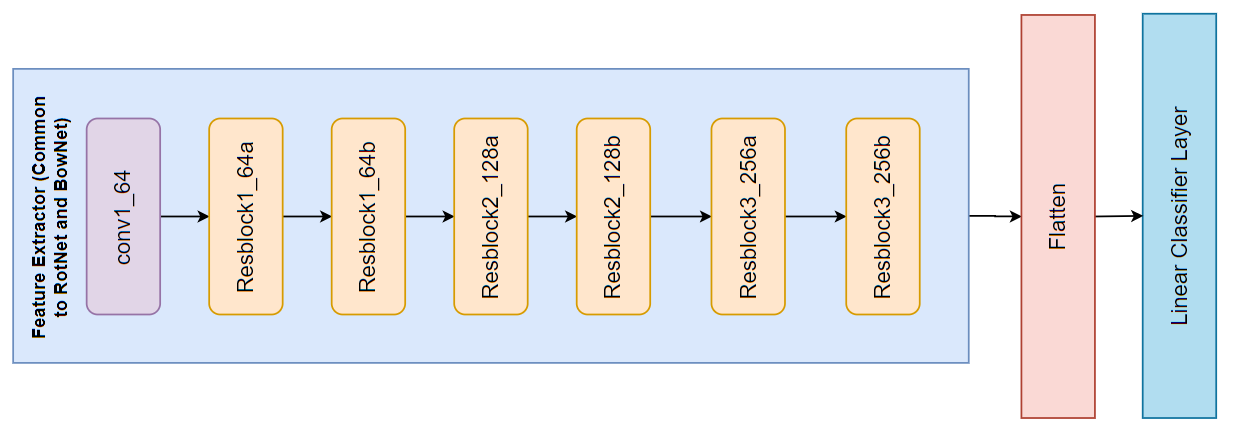}
\caption{Supervised training setup for CIFAR-100 classification. A linear classifier layer receives the flattened feature maps of the ResBlock3\_256b block from either RotNet or BowNet. We experimented with using Global Average Pooling instead of Flatten prior to the linear classifier and found that flattening the feature maps worked better. This is likely due to the fairly low channel depth of the feature map. Global Average Pooling would further discard the relatively sparse amount of information in the feature maps.}
\label{fig:cifar}
\end{figure*}

There were mainly three different training setups we had, with the architecture of the feature extractor common to all three. There was the rotation prediction training, BoW representation training, and supervised CIFAR-100 training. See Figures~\ref{fig:rotnet},~\ref{fig:bownet},~\ref{fig:cifar} for detailed description of each CNN architecture. It is worth noting that the authors introduced a reparametrized linear-plus-softmax layer which we have reproduced to the best of our understanding. However, without reference code, this could be a potential source of mismatch between our implementation and that of the authors. Every convolution layer was followed by BatchNorm and Relu with the exception of the $1\times1$ convolutions that were used in residual blocks for projecting the residual into the same dimension as the Conv feature map that it would be added to in the elementwise add. Due to our hardware limitations, we used a much smaller CNN architecture. To illustrate, WRN-28-10 CNN used in \cite{gidaris2020bownet} has several Conv layers of depth 320 and 640. The largest layers we ever use are the Conv512 residual layers that were used in the rotation prediction task.

\subsection{Experimental setup and code}
Our PyTorch implementation of this method can be found on Github.
Due to the extensive time required to complete an experiment, choices of values to try for hyperparameters such as learning rate, batch size and step size, are estimated a priori instead of performing an extensive hyperparameter search. For instance, the choice of learning rate must be reasonable as it dictates the efficient convergence of a gradient while avoiding local minima, exploding exponentially, or proceed to slowly. Similarly, batch sizes also affect the rate of gradient descent, where a small batch size can reduce convergence time but is prone to being caught in an unintended minima.

For RotNet, we use batch size of $128$ and SGD optimiser with learning rate $0.1$, momentum $0.9$, and weight decay $1\times10{^-6}$. The learning rate will reduce when testing loss had plateaued for 10 epochs. We applied the same learning rate and learning schedule to train the RotNet pretrained + Linear classifier.

For BowNet, we used an initial learning rate of 0.01 and batch-size of 128. For learning rate schedule, we would reduce learning rate when testing loss had plateaued for 10 epochs. We had initially used validation loss, but as we ran into issues with reproducing the author's results we started directly using test loss to see if we could make any gains in accuracy. SGD optimizer with momentum of 0.9 and weight decay of $5\times10^{-4}$ was used for weight updates.

We use accuracy (number of correct predictions/total predictions) to evaluate the experiments.

\subsection{Datasets}
The dataset we used is CIFAR-100 due to its small memory footprint compared to MiniImageNet and ImageNet. CIFAR-100 has 50000 training samples and 10000 testing samples. The labels are categorised in 100 classes and the classes are evenly distributed. The dataset can be downloaded from here \href{https://www.cs.toronto.edu/~kriz/cifar.html}{CIFAR-100}\footnote{\url{https://www.cs.toronto.edu/~kriz/cifar.html}}.\\
\\
To train RotNet, we rotate each image sample 3 times to create $90 \degree, 180 \degree,270 \degree$ rotated images. The original image $0 \degree$ and $90 \degree, 180 \degree,270 \degree$ rotated images are used as labels.\\
\\To train RotNet + nonlinear/linear classifier, we perform several data augmentation techniques over the training set such as color jitter, random crop, random grey color, random horizontal flip to get more data. The labels are 100 classes.\\
\\To train BowNet, we apply a series of five perturbations that are all available as built-in PyTorch functions. These functions, along with their input arguments are: ColorJitter (brightness = 0.2, contrast = 0.2, saturation = 0.2, hue = 0.2), RandomResizeCrop (size = 32, scale = (0.8, 1.0), ratio=(0.75, 1.33), interpolation = 2), RandomCrop (size = 32, padding = 4, padding\_mode = `reflect'), RandomHorizontalFlip (p = 0.5), RandomGreyscale (p = 0.3)). BowNet is trained to take the perturbed images as input and reproduce the same BoW representation as the unperturbed image.

\subsection{Computational requirements}
We performed our experiments using an Intel® Core™ i7-7700 CPU @ 3.60GHz, 32Gb RAM and GTX 1080Ti graphic card.\\

Training RotNet for 200 epoches took 10 hours to complete. We used a batch size of 128 unique images. Since we included copies of all four rotations for each unique image, the effective batch-size being processed by RotNet was 512 on each forward pass.\\

Training a linear classifier with RotNet pretrained as a feature extractor for 200 epoches  took 2 hours to complete.Training a linear classifier with RotNet feature extractor from scratch for 200 epoches took 4 hours to complete. Thus, reproducing and tuning our RotNet implementation required significant time as we ran multiple experiments to validate the behaviour of RotNet. The experimental setup for BowNet followed a similar structure. 
Applying MiniBatchKMeans clustering took 30 minutes to complete (note that this was performed on a CPU using Scikit-learn).\\

Training BowNet to reproduce BoW representations for 30 epochs took 2 hours to complete, and finally, training a linear classifier with BowNet feature extractor for 200 epochs took 2 hours to complete.

\section{Results}

Overall, we were not able to reproduce the BowNet prediction and our CIFAR-100 accuracy using BowNet and a linear classifier (47.20\% testing accuracy) were much worse than training directly with RotNet features (55.67\% testing accuracy). Using a smaller network we were able to get almost within 4\% of the RotNet testing accuracy on CIFAR-100 originally reported in~\cite{gidaris2020bownet}. Therefore, we are inclined to believe that we did reproduce the pretrained RotNet correctly. Our current hypothesis is that our pretrained RotNet was not able to learn enough high-level, semantic representations for the Deep Clustering to create a discriminative codebook. It is unclear whether other aspects of our training pipeline also had errors. Given additional time, we believe we would be able to locate the exact source of the problem.

\begin{table}[ht]
{\begin{minipage}{\textwidth}
\centering
\resizebox{0.55\textwidth}{!}{%
\begin{tabular}{l|cc}
\hline 
Experiments & Accuracy  \\
\hline 
RotNet (rotation prediction) &	78.53\% \\
RotNet (pretrained, resblock3\_256b) + linear clf  	&53.26\%\\
RotNet (pretrained, resblock2\_128b) + linear clf  &55.67\%\\
RotNet (pretrained) + nonlinear clf 	&57.44\%\\
RotNet + linear clf (supervised) 	&60.24\%\\
RotNet + nonlinear clf (supervised) 	&66.06\%\\
BowNet (pretrained, resblock3\_256b) + linear clf  	&47.49\%\\
BowNet (pretrained, resblock2\_128b) + linear clf  	&51.10\%\\
BowNet (crossentropy loss)  	&5.31\\
\hline
\textbf{BowNet original (pretrained + linear clf)} \cite{gidaris2020bownet} & 71.5 \%\\
\hline
\end{tabular}}
\caption{Results. The pretrained networks are pretrained with their respective self-supervised learning task (either rotation prediction or BoW reconstruction). For the pretrained networks we also indicate which feature map was fed to the classifier layers. The supervised networks are trained from scratch on CIFAR-100 using the same feature extractor architecture as the pretrained networks. We also report the crossentropy loss that we observed for the BowNet training. It turned out a lot higher than we expected.}
\label{table:result}
\end{minipage}}
\end{table}%
\section{Discussion}

\subsection{What was easy}
The author's main method is incredibly compact and accessible. With the exception of the Reparametrized Linear-Softmax layer, we could implement all of the main parts with basic Numpy, Scikit-learn, and Pytorch functions. We did not have to worry about any complicated, custom layers or complex loss functions. We found that this made the overall code implementation not too complicated. We found the notation and formulation of their BoW representations a bit confusing. It was particularly confusing to figure out how exactly the feature maps of BowNet were fed to the Reparametrized linear layer (Global Average Pooling, Flatten, or some other method) and we were forced to make our best educated guess. A more detailed diagram of the BowNet architecture would have been incredibly helpful for our understanding. Furthermore, familiarity with bag-of-visual words descriptors will greatly help with understanding of this paper. However, given the abundance of libraries available for k-means clustering, and the ubiquity of bag-of-words in computer vision, it would be very straightforward for others in the field to apply this idea of clustering CNN feature maps. While the authors did mention using CutMix augmentation, which would have required some custom processing, they demonstrated strong results even without CutMix. Thus, further reinforcing the accessibility of their idea.

\subsection{What was difficult}
When reviewing the details of RotNet, we had hoped to build off of the previously trained RotNet checkpoints that the authors of \cite{rotnetgithub} had made available online. However, we found that besides a change of CNN architecture, they also did not have any trained checkpoints for the CIFAR-100 dataset. Thus, this became an issue of reproducing \textit{two} experimental setups. Significant time was spent on reproducing RotNet. Each experiment would take at least 10 hours and since we had to retune hyperparameters this translated into a pretty significant time investment. Unfortunately, we could not move ahead with the BowNet training until after we had verified that our RotNet was training correctly. Otherwise it would be difficult to determine whether poor BowNet results were due to an issue with our experimental setup for BowNet training or if they were caused by poorly learned RotNet representations. We outline the challenges we faced with implementing each of these two papers below.

\subsubsection{RotNet}
To verify RotNet, we needed to experiment with a Linear and Non-linear classifier and compare with the results from RotNet \cite{gidaris2018rotnet} as well as those reported in the BowNet \cite{gidaris2020bownet}. Specifically, we froze the trained layers of RotNet and fed the feature maps of Resblock3\_256b to either a linear classifier (hereafter referred to as RotNet+Linear Clf) or a nonlinear classifier (hereafter referred to as RotNet+Nonlinear Clf). By analyzing the general trends in accuracy of our experiments compared to those reported in these two papers, we could have some confidence in our implementation. Ultimately we found that our best RotNet+Linear accuracy was almost within 4\% of that reported by \cite{gidaris2020bownet}. Furthermore, our RotNet+Linear Clf and RotNet+Nonlinear Clf accuracies seemed reasonable. This process of verification involved many experiments with several different hyperparameter settings and data augmentations (for the CIFAR-100 training) tried. There was significant variation in CIFAR-100 testing accuracy depending on the learning rate schedule and data augmentations used. Thus, these details were not trivial. Likely owing to space constraints, we found that there were some missing details on how the actual CIFAR-100 training was setup for certain steps such as the batch-size and learning rate schedules for RotNet pretraining, choice of weight initialization method for all experiments, and any preprocessing of the CIFAR-100 images (eg. normalization, mean-removal, variance scaling etc.). 

We have experimented with different learning rates, learning rate schedules (reducing learning rate based on number of epochs and reducing learning rate if loss plateau), batch sizes (32, 64, 128) and data augmentation scheme to get the best testing accuracy.   
As a further sanity check, we trained the feature extractor + linear classifier from scratch (Ie. fully supervised). This was to match up our own comparisons with those reported in the RotNet \cite{gidaris2018rotnet}. The same hyperparameters tuning processes was applied here. 

This tuning process had to be repeated when we started training BowNet for the supervised classification task.
It was essential that we reproduced this step correctly so as to ensure that the features learned for the BoW codebook would actually be useful.
The best testing accuracy we can achieve with pretrained feature extractor RotNet and a linear classifier is $55.67 \%$. 
The best testing accuracy we can achieve with trained-from-scratch feature extractor RotNet and a linear classifier is $60.24 \%$. 
\subsubsection{BowNet}
As outlined above, we narrowed the scope of reproducibility to just exploring the accuracy gain from using RotNet features versus BowNet features. However, this considerably increased our overhead with experimental design since we could not simply reuse the training hyperparameters and other experimental setup described in the BowNet paper. This meant that while trying to debug our BowNet results, we had to simultaneously verify the correctness of our RotNet implementation as well as verify the correctness of our codebook generation, BoW reconstruction training etc. Our final results were still decent compared to prior SSL results, but it was nowhere near the same gain that was reported in \cite{gidaris2020bownet}. This leads us to believe that perhaps the feature maps that we used for the KMeans clustering did not contain enough semantic information about the input images. This could potentially be explained by the reduced CNN architecture that we used having a significantly smaller channel dimension. Perhaps BowNet requires features embedded in a very high-dimensional space to achieve good accuracy or we should have reduced the number of centroids in our BoW codebook $K$. Conversely, it is also possible that the CNN we used was simply not big enough to learn useful features for clustering, regardless of the size of the channel dimension. Finally, we cannot rule out the possibility that there was simply an error in our implementation.

We found the author's notation somewhat confusing in terms of figuring out which feature maps  are clustered to create the bag-of-words vs which feature \textbf{vectors} were fed to the Reparameterized Linear Softmax layer for BOW reconstruction. They use $\Phi(x)$ to describe both and it became hard to keep track of which quantity it refered to when reading the mathematical formulations and descriptions of the different models involved. 

Given more time, we believe we could figure out the cause of our lackluster BowNet results. It is hard to say whether the smaller architecture is the cause of our failure or some other factor. If we are indeed correct that our reduced architecture learned poor representations, then this suggests potential issues with the generalizability of this work. The potential applications such as few-shot learning and autonomous driving are typically contexts where compact CNNs are required. Thus, if this method does not work well with small CNN architectures then it raises questions about the applicability of adapting BowNet for tasks requiring efficient CNNs.

\subsection{Conclusion}

Though we obtained some reasonable results, we were not able to reproduce the large accuracy gain over RotNet that was reported from using BoW reconstruction to learn robust, high-level features. Given the significant time invested in verifying the correctness of our RotNet implementation, we believe that with more time we could have better verified if this failure to reproduce the paper's results were due to some error in our implementation, or due to the design choices we made.
\subsection{Communication with original authors}
We tried reaching out to the authors via email and got an initial reply pointing us to relevant literature. Unfortunately, when we followed up asking if they would be open to providing feedback for the reproducibility challenge we did not receive a reply.

\bibliographystyle{unsrt}
\bibliography{main.bib}

\end{document}